\documentclass[runningheads]{llncs}
\usepackage[T1]{fontenc}
\usepackage{graphicx}
\usepackage{multirow}
\usepackage{caption}
\usepackage{amsmath}
\usepackage{xcolor,colortbl} 
\usepackage{booktabs}
\usepackage{threeparttable}
\usepackage{array}
\begin{document}
\title{Cross-Domain Transfer and Few-Shot Learning for Personal Identifiable Information Recognition}
%
%
\author{Junhong Ye \and 
Xu Yuan \and
Xinying Qiu\textsuperscript{*} }

%
\authorrunning{Ye et al.}

\institute{Department of Computer Science, School of Information Science and Technology\\
Guangdong University of Foreign Studies, Guangzhou, China\\ \email{xy.qiu@foxmail.com} 
}
%
%
\maketitle              
\begin{abstract}
Accurate recognition of personally identifiable information (PII) is central to automated text anonymization. This paper investigates the effectiveness of cross-domain model transfer, multi-domain data fusion, and sample-efficient learning for PII recognition. Using annotated corpora from healthcare (I2B2), legal (TAB), and biography (Wikipedia), we evaluate models across four dimensions: in-domain performance, cross-domain transferability, fusion, and few-shot learning. Results show legal-domain data transfers well to biographical texts, while medical domains resist incoming transfer. Fusion benefits are domain-specific, and high-quality recognition is achievable with only 10\% of training data in low-specialization domains.

\end{abstract}
\footnotetext[1]{
	\textsuperscript{*}Corresponding author\\
}

\section{Introduction}
Privacy protection has become a cornerstone of modern data governance, with regulations like the European Union's General Data Protection Regulation (GDPR) \cite{gdpr2016} establishing strict requirements for handling personally identifiable information (PII). As organizations increasingly rely on text-based data for research, analytics, and machine learning, the need for effective text anonymization has grown correspondingly urgent. Traditional manual anonymization approaches are costly, inefficient, and prone to human error \cite{bier2009}, driving the development of automated PII recognition systems using Natural Language Processing (NLP) techniques \cite{meystre2010} \cite{sanchez2016} \cite{dernoncourt2017}.

However, three critical gaps remain unaddressed. First, the cross-domain transferability of PII recognition models remains largely uncharacterized \cite{papadopoulou2022}. Second, organizations often possess data from multiple domains, yet the potential benefits or drawbacks of combining heterogeneous training data for PII recognition remain unclear. Third, many practical deployment scenarios involve limited training data, but minimum data requirements for effective privacy protection have not been systematically evaluated.

To address these gaps, this paper conducts a comprehensive empirical study across healthcare (I2B2), legal (TAB), and biographical (Wikipedia) domains. Our systematic evaluation across 231 experimental configurations yields the following contributions:
\begin{itemize}
\item\textbf{Cross-Domain Transfer Characterization}: We systematically analyze transfer performance across six domain pairs, revealing domain complexity hierarchies and identifying which domains serve as effective transfer sources.
\item\textbf{Multi-Domain Fusion Analysis}: We demonstrate that fusion benefits vary dramatically by target domain, providing domain-specific strategies for optimal performance.
\item\textbf{Sample Efficiency Validation}: We establish that effective PII recognition is achievable with significantly reduced training data, revealing non-monotonic relationships between data size and performance.
\end{itemize}

These findings challenge conventional assumptions about data requirements. We provide our codes at https://github.com/George-SGY/multi-domain-pii-recognition.

\section{Related Research}
Text anonymization research has evolved from rule-based approaches to sophisticated neural models, with most work focusing on domain-specific solutions rather than cross-domain applications that motivate our investigation.

\textbf{Domain-Specific PII Recognition}: 
The medical domain has served as the primary testbed for automated PII recognition, benefiting from well-established datasets like I2B2 \cite{stubbs2015} and achieving remarkable performance with transformer-based approaches such as SciBERT \cite{velupillai2009} and BioBERT \cite{alfalahi2012} reaching F1 scores exceeding 0.98 \cite{marimon2019}. Legal domain research has been more limited, with Pilán et al. \cite{pilan2022} introducing the Text Anonymization Benchmark (TAB) for GDPR-compliant legal document anonymization. Other domains have seen sporadic development, including email communication datasets \cite{keila2005} \cite{medlock2006} and dialogue corpora \cite{patel2013} \cite{jensen2021}.

\textbf{Cross-Domain and Multi-Domain Approaches}: 
Despite the proliferation of domain-specific datasets and models, systematic cross-domain evaluation remains scarce. Papadopoulou et al. \cite{papadopoulou2022} made a notable contribution with a Wikipedia-based biography corpus and distant supervision methods, but their work primarily focused on reducing annotation requirements rather than systematic transfer analysis. The broader NLP literature has extensively studied domain adaptation for tasks such as sentiment analysis and named entity recognition \cite{bier2009}, establishing transfer learning as effective for leveraging knowledge across domains. However, these insights have not been systematically applied to PII recognition, where domain-specific privacy requirements and entity distributions may pose unique challenges.

\section{Methodology}
This section presents our comprehensive experimental framework for investigating cross-domain PII recognition. We systematically evaluate model performance across multiple domains, training paradigms, and data availability scenarios to address the four key research questions outlined in the introduction. Overall research framework is presented in Figure 1.
\begin{figure}
\centering
\includegraphics[scale=0.58]{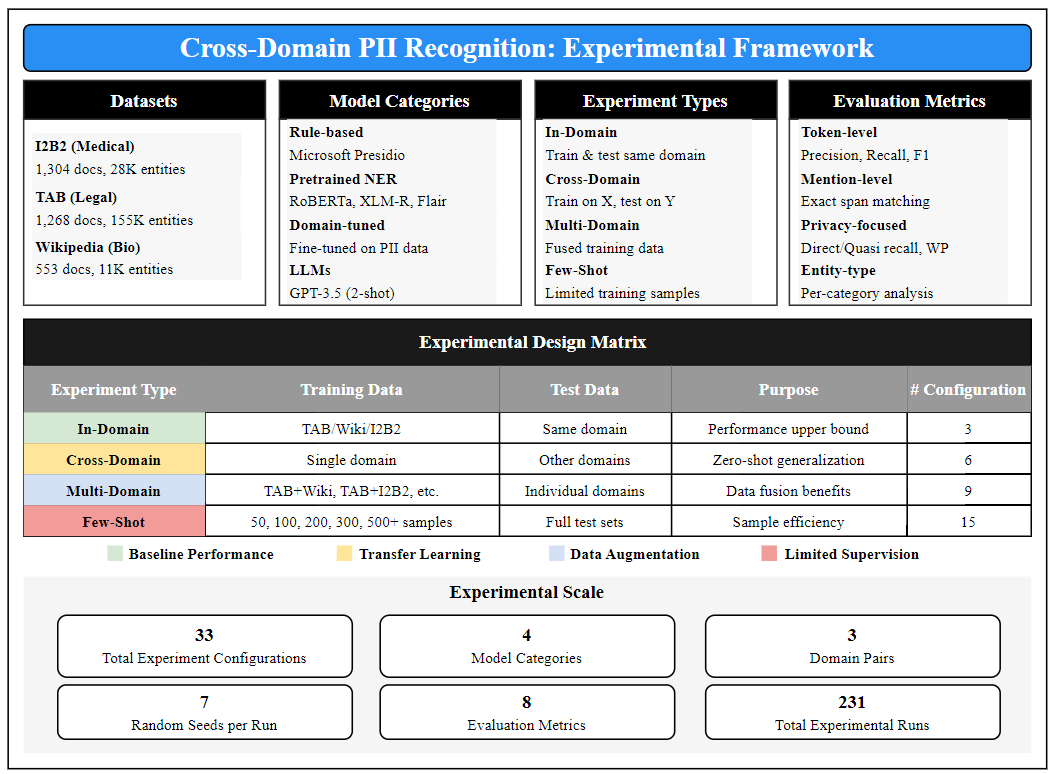}
\caption{Research Framework} 
\label{fig1}
\end{figure}

\subsection{Datasets and Data Preparation}
We selected three publicly available, human-annotated datasets representing distinct domains with varying structural characteristics and annotation schemes:
\begin{itemize}
\item \textbf{I2B2 (Medical Domain)}: 1,304 clinical narratives with 28,867 entity mentions across 23 fine-grained medical entity types. Contains 790 training and 514 test documents (average: 925 tokens).
\item \textbf{TAB (Legal Domain)}: 1,268 legal documents with 155,006 entity mentions across 8 semantic types (PERSON, DATETIME, ORG, LOC, DEM, MISC, CODE, QUANTITY). Contains 1,014 training, 127 validation, and 127 test documents (average: 1,442 tokens).
\item \textbf{Wikipedia (Biographical Domain)}: 553 Wikipedia biography entries with 10,714 entity mentions across 7 semantic types. Contains 453 training and 100 test documents (average: 96 tokens).
\end{itemize}

All datasets distinguish between direct identifiers (entities that uniquely identify individuals) and quasi-identifiers (contextual attributes that can reveal identity when combined). Text tokenization was performed using SpaCy with consistent handling across domains, and entity spans were converted to BIO tagging format.

\subsection{Model Architectures}
We benchmark four distinct model categories, each representing different approaches to PII recognition:
\begin{itemize}
\item \textbf{Rule-based Systems}: Microsoft Presidio applying regex patterns and heuristic-based entity detection
\item \textbf{Pretrained NER Models}: RoBERTa-large, XLM-RoBERTa, and Flair NER models pretrained on OntoNotes 5.0, applied zero-shot without domain-specific fine-tuning
\item \textbf{Domain-Tuned NER Models}: The same transformer architectures (RoBERTa, Longformer, Flair) fine-tuned on domain-specific PII data. Longformer handles longer documents with its extended attention mechanism (4096 vs. 512 tokens).
\item \textbf{Large Language Models (LLMs)}: GPT-3.5-turbo evaluated under 2-shot in-context learning.
\end{itemize}

Based on preliminary experiments, we selected Longformer as our primary model for cross-domain experiments due to its consistent performance and ability to handle varying document lengths.

\subsection{Experimental Design and Evaluation}
We conduct four complementary experiments :
\begin{itemize}
\item\textbf{In-Domain Evaluation}: Models trained and tested within the same domain, establishing performance upper bounds and domain-specific baselines.
\item\textbf{Cross-Domain Transfer}: Models trained on one domain's training set and evaluated on other domains' test sets, assessing zero-shot generalization capabilities across the 3 by 3 domain matrix.
\item\textbf{Multi-Domain Fusion}: Models trained on merged corpora from multiple domains (e.g., TAB+Wikipedia, TAB+I2B2, all three domains) and evaluated on individual domain test sets, investigating whether diverse training data improves domain-specific performance.
\item\textbf{Few-Shot Learning}: Systematic reduction of training data volume (50, 100, 200, 300, 500+ samples) to evaluate model robustness under limited supervision and validate practical deployment scenarios.
\end{itemize}
We use token-level evaluation metrics of precision, recall, and F1. All experiments are conducted across seven independent runs with different random seeds. We report median scores for robust performance estimates.

\section{Results and Analysis}
This section presents our experimental findings across four dimensions: in-domain performance, cross-domain transferability, multi-domain data fusion, and few-shot learning feasibility.

\subsection{Baseline Performance Analysis}
Table 1 presents consolidated results across all model categories and domains. Domain-tuned transformer models (Longformer, RoBERTa, Flair) consistently outperform other approaches, achieving F1 scores of 0.83-0.97. Longformer demonstrates the most balanced performance with high recall while maintaining reasonable precision. Rule-based approaches (Presidio) show the weakest performance (F1: 0.57-0.65). Pretrained NER models suffer from low and highly varied precisions (0.32-0.799). GPT-3.5-turbo demonstrates moderate performance (F1: 0.63-0.868) with superior precision compared to pretrained NER models. NER Fine-Tuned models perform second best overall after Domain-Tuned models.

Among domain-tuned models, Longformer achieves the best performances on all three evaluation metrics for medical domain (I2B2) and legal domain (TAB) and best recall for biographical domain (Wikipedia). Based on these results, we select Longformer for subsequent cross-domain experiments.

\begin{table*}[htbp]
\centering
\caption{Cross-Domain Model Performance Summary. Results show F1/Recall/Precision for token-level evaluation across three domains. Best results per category are in \textbf{bold}.}
\label{tab:model_performance_summary}
\resizebox{\textwidth}{!}{%
\begin{tabular}{@{}llccccc@{}}
\toprule
\textbf{Model Category} & \textbf{Model} & \textbf{TAB} & \textbf{Wikipedia} & \textbf{I2B2} & \textbf{Average F1} & \textbf{Rank} \\
 & & \textit{F1/Rec/Prec} & \textit{F1/Rec/Prec} & \textit{F1/Rec/Prec} & & \\
\midrule
\multirow{1}{*}{Rule-based} 
& Presidio & 0.649/0.744/0.576 & 0.642/0.558/0.757 & 0.573/0.779/0.453 & 0.621 & 10 \\
\midrule
\multirow{4}{*}{\shortstack{Pretrained\\NER}} 
& RoBERTa & 0.563/0.886/0.413 & 0.833/0.870/0.799 & 0.466/0.838/0.323 & 0.621 & 10 \\
& XLM-RoBERTa & 0.574/0.906/0.420 & 0.821/0.846/0.798 & 0.512/0.723/0.396 & 0.636 & 9 \\
& Flair & 0.560/0.913/0.404 & 0.824/0.866/0.785 & 0.439/0.633/0.336 & 0.608 & 11 \\
& Flair-fast & 0.562/0.910/0.407 & 0.817/0.858/0.780 & 0.416/0.589/0.321 & 0.598 & 12 \\
\midrule
\multirow{3}{*}{\shortstack{Domain-Tuned\\(PII only)}} 
& Longformer & \textbf{0.859/0.937/0.793} & 0.828/\textbf{0.986}/0.713 & \textbf{0.972/0.994/0.950} & \textbf{0.886} & \textbf{1} \\
& RoBERTa & 0.857/0.934/0.791 & 0.834/0.982/0.723 & 0.959/0.937/0.982 & 0.883 & 2 \\
& Flair & 0.873/0.857/\textbf{0.889} & \textbf{0.884/0.875/0.893} & 0.941/0.904/\textbf{0.982} & 0.866 & 5 \\
\midrule
\multirow{5}{*}{\shortstack{NER+\\Fine-Tuned}} 
& RoBERTa & 0.826/0.943/0.756 & 0.841/0.980/0.737 & 0.962/0.987/0.939 & 0.876 & 4 \\
& XLM-RoBERTa & 0.852/0.947/0.784 & 0.824/0.978/0.712 & 0.965/0.989/0.945 & 0.880 & 3 \\
& Flair & 0.869/0.856/0.882 & 0.877/0.866/0.888 & 0.948/0.913/0.985 & 0.865 & 6 \\
& Flair-fast & 0.865/0.833/0.899 & 0.874/0.861/0.888 & 0.941/0.901/0.985 & 0.860 & 7 \\
\midrule
\multirow{1}{*}{LLM} 
& GPT-3.5-turbo & 0.738/0.728/0.748 & 0.868/0.907/0.833 & 0.630/0.765/0.535 & 0.745 & 8 \\
\bottomrule
\end{tabular}%
}
\begin{tablenotes}
\small
\item Results averaged across 7 random seeds. 
\item Domain-tuned models consistently outperform other categories, with Longformer achieving the best overall performance.
\end{tablenotes}
\end{table*}

\subsection{Cross-Domain Transfer Analysis}
Table 2 summarizes transfer performance across all domain pairs. 

\textbf{TAB $\rightarrow$ Other Domains}: Legal domain training demonstrates excellent transferability to biographical content (Wikipedia F1=0.847, +2.3\% improvement) while showing severe degradation for medical domains (I2B2 F1=0.264, -72.8\% drop). 

\textbf{I2B2 $\rightarrow$ Other Domains}: Medical training shows surprisingly effective transfer, maintaining baseline performance on Wikipedia (F1=0.828, 0\% change) and achieving reasonable performance on TAB (F1=0.738, -14.1\% drop). 

\textbf{Wikipedia $\rightarrow$ Other Domains}: Biographical training exhibits variable performance, achieving moderate results on TAB (F1=0.421, -51\% drop) but failing dramatically on medical data (I2B2 F1=0.187, -80.8\% drop).

\textbf{Key Insights}: (1) Transfer target hierarchy: I2B2 is most difficult to transfer into, while Wikipedia is most receptive; (2) TAB provides the most effective transfer supervision for biographical domains; (3) Medical domains resist incoming transfer but provide surprisingly effective outgoing transfer.

\begin{table*}
\centering
\caption{Cross-Domain Transfer Performance using Longformer (Token F1 scores). Diagonal entries show in-domain baselines. Best cross-domain transfers per target are in \textbf{bold}.}
\label{tab:cross_domain_transfer}
\begin{tabular}{@{}lccc@{}}
\toprule
\textbf{Train $\backslash$ Test} & \textbf{TAB} & \textbf{Wikipedia} & \textbf{I2B2} \\
\midrule
TAB & \cellcolor{lightgray}0.859 & \textbf{0.847} & 0.264 \\
Wikipedia & \textbf{0.421} & \cellcolor{lightgray}0.828 & 0.187 \\
I2B2 & 0.738 & 0.828 & \cellcolor{lightgray}0.972 \\
\midrule
\textbf{Performance Change} & \textbf{-32.5\%} & \textbf{+1.15\%} & \textbf{-76.8\%} \\
\bottomrule
\end{tabular}

\vspace{0.3cm}

\begin{tabular}{@{}lcccc@{}}
\toprule
\textbf{Transfer Direction} & \textbf{F1} & \textbf{Recall} & \textbf{Precision} & \textbf{vs Baseline} \\
\midrule
\multicolumn{5}{l}{\textit{Successful Transfers (F1 > 0.7):}} \\
TAB $\rightarrow$ Wikipedia & 0.847 & 0.926 & 0.781 & +2.3\% \\
I2B2 $\rightarrow$ Wikipedia & 0.828 & 0.803 & 0.825 & 0\% \\
I2B2 $\rightarrow$ TAB & 0.738 & 0.840 & 0.658 & -14.1\% \\
\midrule
\multicolumn{5}{l}{\textit{Challenging Transfers (F1 < 0.5):}} \\
Wikipedia $\rightarrow$ TAB & 0.421 & 0.991 & 0.267 & -51\% \\
TAB $\rightarrow$ I2B2 & 0.264 & 0.990 & 0.152 & -72.8\% \\
Wikipedia $\rightarrow$ I2B2 & 0.187 & 0.999 & 0.103 & -80.8\% \\
\bottomrule
\end{tabular}

\begin{tablenotes}
\small
\item Results averaged across 7 random seeds using Longformer architecture.
\item Gray cells show in-domain baselines. Performance drops show average degradation when transferring into each target domain.
\end{tablenotes}
\end{table*}

\subsection{Multi-Domain Fusion Effects}
Table 3 presents fusion results across all domain combinations.

\textbf{TAB Domain} benefits from multi-domain fusion, with the three-domain fusion (TWI) achieving the highest recall (0.952, +1.6\% over baseline) despite a modest decrease in F1 (-0.7\%). This indicates that fusion enhances privacy protection by reducing the likelihood of missed PII, even if overall balance slightly declines.
 
\textbf{Wikipedia Domain} shows nuanced responses. Adding I2B2 data (WI) raises precision (+3.1\%) with negligible recall change, while combining all three domains (TWI) yields the best overall F1 (+2.5\%). This suggests that the optimal strategy depends on whether precision or overall balance is prioritized, though TWI offers the strongest aggregate performance.
 
\textbf{I2B2 Domain} fusion consistently harms performance across all metrics, with F1 decreasing by 0.2–0.5\% and no recall improvements compared to baseline. This reinforces the medical domain’s high specialization and indicates that domain-specific training is the only reliable strategy.

\textbf{Strategy Implications}: Legal documents benefit from fusion due to improved recall, biographical texts gain most from full-domain fusion, while medical records remain best served by single-domain training only.

\begin{table}[htbp]
\centering
\scriptsize
\caption{Multi-Domain Data Fusion Effects on Target Domain Performance. Results show F1/Recall/Precision for token-level evaluation using Longformer. Best fusion results per domain are in \textbf{bold}.}
\label{tab:fusion_results}
\begin{tabular}{@{}lccccc@{}}
\toprule
\textbf{Target} & \textbf{Training Data} & \textbf{F1} & \textbf{Recall} & \textbf{Precision} & \textbf{vs Baseline F1} \\
\midrule
\multirow{4}{*}{\textbf{TAB}} 
& TAB only (baseline) & 0.859 & 0.937 & 0.793 & --- \\
& TAB + Wiki (TW) & 0.841 & 0.946 & 0.757 & -2.1\% \\
& TAB + I2B2 (TI) & 0.841 & 0.945 & 0.757 & -2.1\% \\
& TAB + Wiki + I2B2 (TWI) & \textbf{0.853} & \textbf{0.952} & \textbf{0.772} & \textbf{-0.7\%} \\
\midrule
\multirow{4}{*}{\textbf{Wiki}} 
& Wiki only (baseline) & 0.828 & 0.986 & 0.713 & --- \\
& Wiki + TAB (TW) & 0.820 & \textbf{0.989} & 0.700 & -1.0\% \\
& Wiki + I2B2 (WI) & 0.842 & 0.985 & 0.735 & +1.7\% \\
& Wiki + TAB + I2B2 (TWI) & \textbf{0.849} & 0.973 & \textbf{0.753} & \textbf{+2.5}\% \\
\midrule
\multirow{4}{*}{\textbf{I2B2}} 
& I2B2 only (baseline) & 0.972 & 0.994 & 0.950 & --- \\
& I2B2 + TAB (TI) & 0.968 & \textbf{0.994} & 0.944 & -0.4\% \\
& I2B2 + Wiki (WI) & \textbf{0.970} & 0.993 & \textbf{0.948} & \textbf{-0.2}\% \\
& I2B2 + TAB + Wiki (TWI) & 0.967 & 0.991 & 0.944 & -0.5\% \\
\bottomrule
\end{tabular}

\vspace{0.3cm}

\begin{tabular}{@{}lcccc@{}}
\toprule
\textbf{Domain Characteristics} & \textbf{Best Strategy} & \textbf{F1 Effect} & \textbf{Recall Effect} & \textbf{Recommendation} \\
\midrule
\textbf{TAB (Legal)} & All domains (TWI) &  -0.7\% & +1.6\% & Fusion beneficial \\
\textbf{Wiki (Biographical)} & All domains (TWI) &  +2.5\% & -1.3\% & Fusion beneficial \\
\textbf{I2B2 (Medical)} & Domain-specific  & -0.2\% to -0.5\% & -0.3\% & Avoid fusion \\
\bottomrule
\end{tabular}

\begin{tablenotes}
\scriptsize
\item Fusion combinations: TW (TAB+Wikipedia), TI (TAB+I2B2), WI (Wikipedia+I2B2), TWI (all three domains).
\item Results averaged across 7 random seeds. Positive percentages indicate improvement over single-domain baseline.
\end{tablenotes}
\end{table}

\subsection{Few-Shot Learning Feasibility}
Table 4 shows performance across varying training set sizes. \textbf{TAB Domain} exhibits a non-monotonic pattern, with peak performance observed at 300 samples (-0.8\% F1 vs. full data). Even with only 50 samples, the model retains 97\% of full-data performance (-2.9\% F1), indicating strong robustness under limited supervision.

\textbf{Wikipedia Domain} demonstrates particularly low sensitivity to training size. Using 200 samples (44\% of the full set) yields the best F1 (+2.1\%), while reductions to 50 or 100 samples still maintain performance within 2\% of the full baseline. This suggests that additional training data may introduce noise or complexity rather than improving recognition quality.

\textbf{I2B2 Domain} shows a monotonic relationship with data volume. With 50 samples, F1 drops by 6.6\%, and performance improves steadily as data increases, requiring nearly the full dataset to reach peak accuracy (F1 = 0.972). This reflects the domain’s high complexity and greater dependence on large-scale supervision.

Overall, these results confirm that effective PII recognition is feasible with substantially reduced data in less specialized domains (TAB, Wikipedia), while medical records (I2B2) demand larger datasets to achieve comparable performance.

\begin{table}[htbp]
\centering
\scriptsize
\caption{Few-Shot Learning Performance using Longformer. Results show F1/Recall/Precision for token-level evaluation across varying training set sizes. Best performance per domain is in \textbf{bold}.}
\label{tab:fewshot_results}
\begin{tabular}{@{}lcccccc@{}}
\toprule
\textbf{Domain} & \textbf{Training Size} & \textbf{F1} & \textbf{Recall} & \textbf{Precision} & \textbf{vs Full Data} & \textbf{Samples/Entity} \\
\midrule
\multirow{5}{*}{\textbf{TAB}} 
& 50 samples & 0.834 & 0.940 & 0.770 & -2.9\% & 0.03 \\
& 300 samples & \textbf{0.852} & \textbf{0.947} & 0.774 & \textbf{-0.8\%} & 0.19 \\
& 550 samples & 0.849 & 0.945 & 0.776 & -1.2\% & 0.35 \\
& 800 samples & 0.848 & 0.934 & 0.775 & -1.3\% & 0.52 \\
& 1014 samples (full) & 0.859 & 0.937 & \textbf{0.793} & --- & 0.65 \\
\midrule
\multirow{5}{*}{\textbf{Wiki}} 
& 50 samples & 0.815 & 0.975 & 0.702 & -1.6\% & 0.47 \\
& 100 samples & 0.819 & 0.979 & 0.704 & -1.1\% & 0.93 \\
& 200 samples & \textbf{0.845} & 0.972 & \textbf{0.754} & \textbf{+2.1\%} & 1.87 \\
& 300 samples & 0.836 & 0.982 & 0.728 & +1.0\% & 2.80 \\
& 453 samples (full) & 0.828 & \textbf{0.986} & 0.713 & --- & 4.23 \\
\midrule
\multirow{5}{*}{\textbf{I2B2}} 
& 50 samples & 0.908 & 0.982 & 0.844 & -6.6\% & 0.17 \\
& 100 samples & 0.939 & 0.983 & 0.898 & -3.4\% & 0.35 \\
& 300 samples & 0.966 & 0.991 & 0.942 & -0.6\% & 1.04 \\
& 500 samples & 0.962 & 0.994 & 0.932 & -1.0\% & 1.73 \\
& 790 samples (full) & \textbf{0.972} & \textbf{0.994} & \textbf{0.950} & --- & 2.74 \\
\bottomrule
\end{tabular}

\vspace{0.3cm}

\begin{tabular}{@{}lccc@{}}
\toprule
\textbf{Domain} & \textbf{Optimal Size} & \textbf{Performance Pattern} &  \textbf{Data Sensitivity} \\
\midrule
\textbf{TAB} & ~30\% (300 samples) & Non-monotonic & Low \\
\textbf{Wikipedia} & ~44\% (200 samples) & Peak at mid-scale &  Very Low \\
\textbf{I2B2} & 100\% (790 samples) & Monotonic increase & High \\
\bottomrule
\end{tabular}

\begin{tablenotes}
\small
\item Results averaged across 7 random seeds. Samples/Entity shows training examples per entity type for context.
\item Optimal size represents training data volume that achieves best performance for each domain.
\end{tablenotes}
\end{table}

\subsection{Summary of Key Findings}
Our comprehensive evaluation yields several critical insights:
\begin{enumerate}
\item\textbf{Model Selection}: Domain-tuned transformers, particularly Longformer, provide optimal privacy-utility balance; 
\item\textbf{Cross-Domain Transfer}: Effectiveness varies dramatically by domain pair, with TAB$\rightarrow$Wikipedia achieving excellent results (+2.3\%) while transfers into medical domains remain challenging (73-81\% drops); 
\item\textbf{Data Fusion Strategy}: Fusion effects are domain-specific. Legal documents benefit from all-domain fusion (TWI), which improves recall even with a small F1 trade-off. Biographical texts achieve the strongest overall performance with TWI (+2.5\% F1), while medical records consistently require domain-specific training due to performance degradation under fusion.
\item\textbf{Sample Efficiency}: PII recognition is surprisingly robust to data limitations. Acceptable performance is achieved with substantially reduced data in less specialized domains (TAB and Wikipedia), whereas I2B2 demands nearly full training data due to its higher complexity.
\end{enumerate}

\section{Conclusions}
This paper presents the first large-scale empirical study of cross-domain PII recognition, evaluating model transferability, data fusion strategies, and sample efficiency across healthcare, legal, and biographical texts. Domain-tuned transformers, particularly Longformer, consistently outperform rule-based and general NER models. Cross-domain transfer is highly asymmetric: legal-domain models transfer well to biographical data, but all models perform poorly on medical text. Multi-domain fusion offers gains for legal and biographical domains when all domains are combined, but consistently harms performance on medical data. Effective PII recognition is also achievable with as little as 10\% of training data in non-specialized domains.

These findings suggest legal-domain models may be repurposed for biographical texts with minimal loss, while medical applications require domain-specific training. While results are strong across many settings, future work should address limitations in extending evaluation to conversational and social media data, improve fusion strategies, and explore synthetic data for privacy-preserving few-shot learning.

%
%
%
%

\end{document}